\documentclass{article}
\usepackage{booktabs}
\usepackage{multirow}

\usepackage{amsfonts,spconf}
\usepackage{amsmath,graphicx}
\usepackage{xcolor}
\usepackage{hyperref}
\usepackage{caption}

\title{YOLO-MED : Multi-Task Interaction Network for Biomedical Images}

\name{Suizhi Huang$^1$, Shalayiding Sirejiding$^1$, Yuxiang Lu$^1$, Yue Ding$^1$, Leheng Liu$^2$, Hui Zhou$^{2,*}$, Hongtao Lu$^{1,*}$ \thanks{This paper is supported by National Nature Science Foundation of China (62176155), Shanghai Municipal Science and Technology Major Project (2021SHZDZX0102). Hongtao Lu is also with MOE Key Lab of Artificial Intelligence, AI Institute, Shanghai Jiao Tong University.}}
\address{$^1$ Department of Computer Science and Engineering, Shanghai Jiao Tong University \\ $^2$ Department of Gastroenterology, Shanghai General Hospital}

\begin{document}
\ninept
\maketitle
\begin{abstract}
Object detection and semantic segmentation are pivotal components in biomedical image analysis. Current single-task networks exhibit promising outcomes in both detection and segmentation tasks. Multi-task networks have gained prominence due to their capability to simultaneously tackle segmentation and detection tasks, while also accelerating the segmentation inference. Nevertheless, recent multi-task networks confront distinct limitations such as the difficulty in striking a balance between accuracy and inference speed. Additionally, they often overlook the integration of cross-scale features, which is especially important for biomedical image analysis. In this study, we propose an efficient end-to-end multi-task network capable of concurrently performing object detection and semantic segmentation called YOLO-Med. Our model employs a backbone and a neck for multi-scale feature extraction, complemented by the inclusion of two task-specific decoders. A cross-scale task-interaction module is employed in order to facilitate information fusion between various tasks. Our model exhibits promising results in balancing accuracy and speed when evaluated on the Kvasir-seg dataset and a private biomedical image dataset.

\end{abstract}
\vspace{-5pt}
\section{Introduction}
\label{sec:intro}
\vspace{-3pt}

\begin{figure}[t]
\setlength{\abovecaptionskip}{1pt}
\centering
\includegraphics[width=0.9\linewidth]{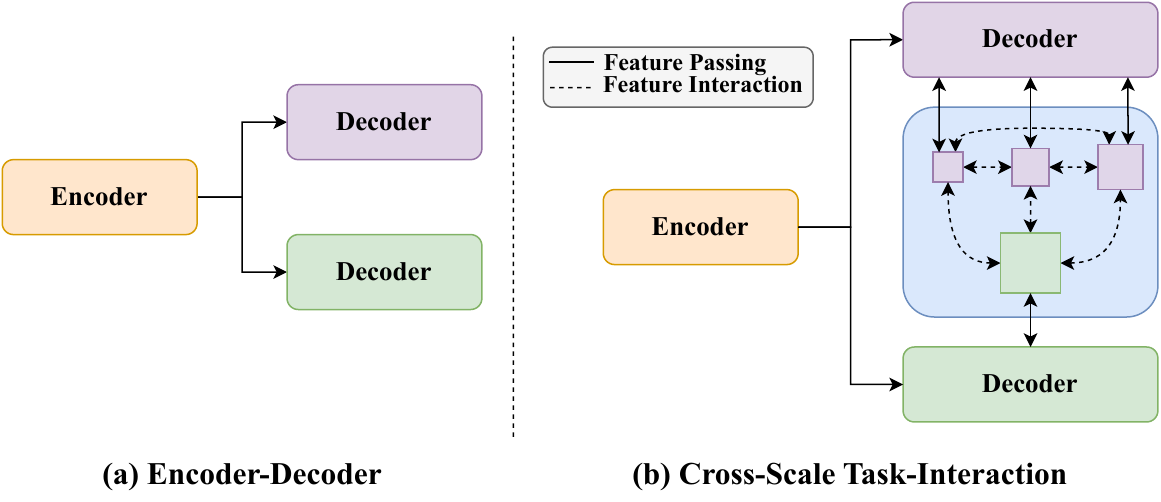}
\caption{Comparison between encoder-decoder structure and our cross-scale task-interaction structure. }
\label{comp}
\vspace{-0.7cm}
\end{figure}

Accurate detection and segmentation of anatomical structures in biomedical images are critical for numerous clinical applications \cite{review3,review4}. Object detection is crucial for identifying abnormalities, like polyps in colonoscopy videos, lesions in retinal fundus images \cite{review1,review2,review5}. Meanwhile, segmentation delineates object boundaries, which facilitates quantitative assessment. For instance, it is widely employed in segmenting polyps, tumor regions, and organs in CT scans \cite{review1,review2,review6}.
Deep learning models have shown immense promise for biomedical image analysis. YOLO series \cite{yolo,yolov4} and RetinaNet \cite{retina} have become classic network architectures in the field of biomedical object detection, while segmentation networks have showcased impressive performance \cite{polyppvt,unet,pranet,progressive,cross}. To address the simultaneous requirements of detection and segmentation \cite{dataset} and the need to accelerate inference, multi-task networks for biomedical image detection and segmentation are employed \cite{uolo,mulan}. Nevertheless, existing multi-task networks for biomedical images still have certain limitations, such as hard to strike a balance between accuracy and inference speed and not adequately taking the use of features from different tasks. Representative networks like UOLO incorporates U-Net as its core and connect it with a YOLO detection head \cite{uolo}, it remains an encoder-decoder architecture like the structure shown in Fig. \ref{comp}, and exclusively relies on U-Net \cite{unet} for extracting shared features across tasks. MULAN adopts a similar shared encoder, and still faces challenges in effectively fusing detection and segmentation features \cite{mulan}. Recently, multi-task networks for dense prediction tasks begin to use inter-task information exchange and achieve significant improvements in accuracy \cite{demt,shala,prompt}. However, these networks are tailored for natural images, which diverge from the unique characteristics of biomedical images. Objects to be detected and segmented in biomedical images usually consist of abnormal cellular tissues that closely resemble the background. Consequently, the incorporation of multi-scale semantic information becomes important in biomedical image analysis. Regrettably, existing networks do not make use of the fusion of cross-scale features.

To address these challenges, we present a novel end-to-end multi-task network for biomedical detection and segmentation. 
We use a backbone to extract a universal representation of input images, then a neck is used to fuse the multi-scale features generated by the backbone. Two task-specific decoders are used to handle segmentation and detection tasks, where unlike traditional approaches, we split the detection tasks (classification and regression) into different branches to improve the detection accuracy. In order to implement the task-interaction, we combine feature maps from segmentation and detection at different scales through a transformer layer, subsequently delivering the fused results to the respective decoder heads.

\begin{figure*}[h]
\centering
\setlength{\abovecaptionskip}{1pt}
\includegraphics[width=0.9\linewidth]{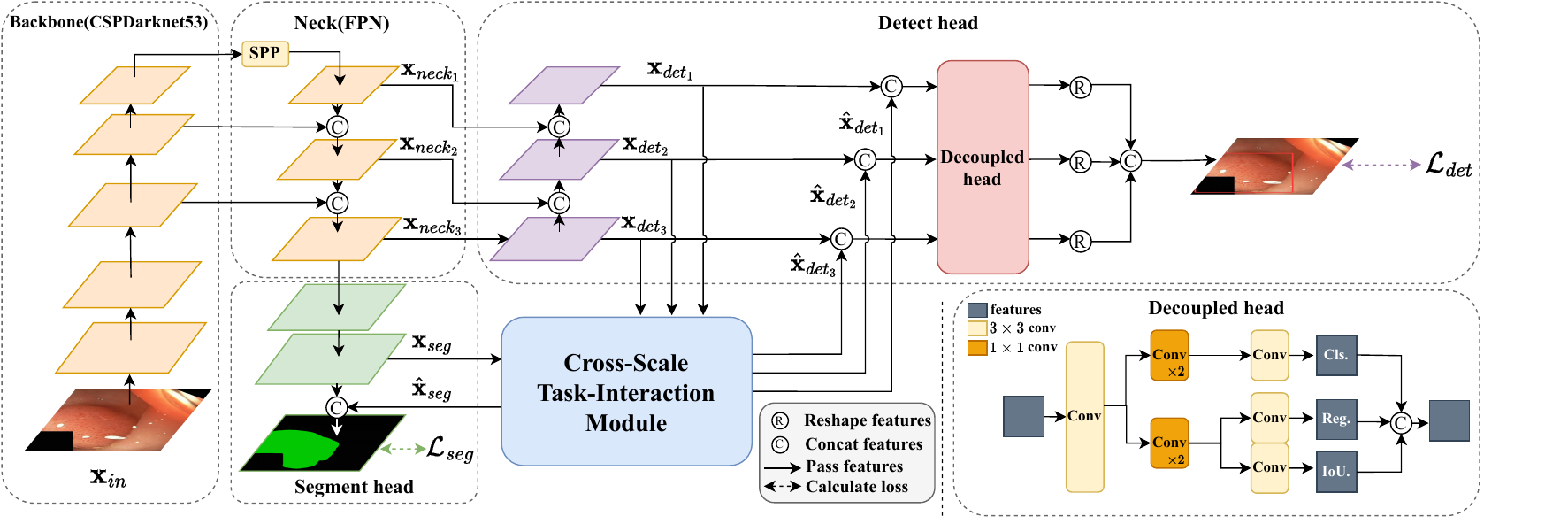}
\caption{The architecture of YOLO-Med network. YOLO-Med shares one encoder and combines 2 decoders with a cross-scale task-interaction module to solve different tasks. The encoder consists of a backbone and a neck, and the detection head has a decoupled head module.}
\label{yolomed}
\vspace{-9pt}
\end{figure*}

In summary, the main contributions of this paper are as follows: 1) We propose YOLO-Med, an efficient end-to-end multi-task network that jointly addresses the tasks of object detection and semantic segmentation in biomedical image analysis. Compared with other multi-task networks for biomedical images, YOLO-Med shows promising results in the trade-off between accuracy and speed.  2) We devise a cross-scale task-interaction module to facilitate interaction between the detection head and segmentation head from multiple scales as shown in Fig. \ref{comp}. Also a decoupled detection head is adopted, which is first time used in multi-task networks for biomedical image detection and segmentation. 3) We validate YOLO-Med on two datasets, Kvasir-seg \cite{Kvasir} and a large private dataset \cite{dataset}. Our results achieve a promising performance across multiple metrics, confirming the effectiveness of YOLO-Med.

\vspace{-8pt}

\section{Method}
\label{sec:method}
\vspace{-3pt}
As shown in Fig. \ref{yolomed}, YOLO-Med consists of a shared encoder and two task-specific decoders, one for each task. Furthermore, the network includes a cross-scale task-interaction module, enabling effective information fusion between the detection and segmentation tasks.

\vspace{-5pt}
\subsection{Encoder}
\vspace{-3pt}
Our network employs a shared encoder consisting of two main components: the Backbone and the Neck.

\noindent \textbf{Backbone.} The backbone network extracts features from input images. We choose CSPDarknet53 \cite{yolov4} as it supports feature propagation and reuse, leading to a significant reduction in parameter and computational overhead during training. This choice guarantees real-time network performance. The initial image data
$\textbf{x}_{in}\in \mathbb{R}^{{H}{\times}W{\times}3}$  is input to the backbone.

\noindent \textbf{Neck.} The Neck is responsible for fusing the multi-scale features generated by the backbone.
First, we pass the output of the backbone through an SPP (Spatial Pyramid Pooling) network \cite{spp} and subsequently feed it into the FPN (Feature Pyramid Network) \cite{fpn}. The SPP module is utilized for feature generation and fusion across multiple scales, while the FPN module combines features from different semantic levels. This fusion process ensures that the resulting features encompass a rich blend of multi-scale and multi-semantic information. Within this module, we obtain three features with different scales: $\textbf{x}_{neck_{i}} \in \mathbb{R}^{\frac{H}{s_i}{\times}\frac{W}{s_i}{\times}c_i}$
where $s_i$ denotes the scale parameter ranging from 8 to 32, and $c_i$ represents the channel number of each feature map ranging from 128 to 512.

\begin{figure}[t]
\setlength{\abovecaptionskip}{1pt}
\centering
\includegraphics[width=0.9\linewidth]{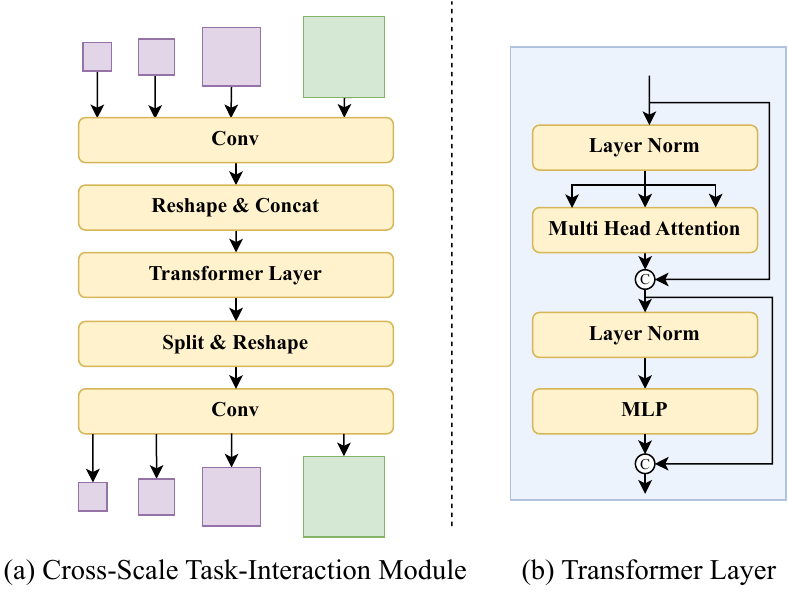}
\caption{The architecture of (a) cross-scale task-interaction module and (b) transformer layer. }
\label{MSTI}
\vspace{-0.5cm}
\end{figure}

\vspace{-5pt}
\subsection{Decoders}
\vspace{-3pt}
In our network, the two heads are specific decoders for detection and segmentation.

\noindent \textbf{Decoupled heads for detection.} First, we construct a Path Aggregation Network (PAN) \cite{pan}. PAN operates as a bottom-up pyramid network, aligning with the top-down semantic propagation in FPN. The diverse scale feature maps $\textbf{x}_{det_i} \in \mathbb{R}^{\frac{H}{s_i}{\times}\frac{W}{s_i}{\times}c_i}$ obtained from PAN are subsequently fused with the correspondingly scaled feature maps $\hat{\textbf{x}}_{det_{i}} \in \mathbb{R}^{\frac{H}{s_i}{\times}\frac{W}{s_i}{\times}c_i}$ generated by the cross-scale task-interaction module. These fused features serve as input to the final detection head. For our final detection head component, we opt for the decoupled head architecture \cite{yolox}. This choice is based on the recognition that traditional coupled heads have demonstrated performance limitations due to inherent conflicts between classification and regression tasks \cite{rethink}. Therefore, within our multi-task architecture, we introduce decoupled heads to ensure that each task (classification and regression) does not negatively affect the others.

\noindent \textbf{Segment head.} For the segmentation head, we devise a straightforward top-down network structure. 
Initially, we take the feature map from the lowest level of the FPN as input. Following two rounds of feature integration and upsampling, the resulting feature map $\textbf{x}_{seg} \in \mathbb{R}^{\frac{H}{2}{\times}\frac{W}{2}{\times}32}$ is merged with the feature map $\hat{\textbf{x}}_{seg} \in \mathbb{R}^{\frac{H}{2}{\times}\frac{W}{2}{\times}32}$ generated by the cross-scale task-interaction module. This fused feature map is subsequently used as input for the final round of feature integration and upsampling. 

\begin{table}[t]
\scriptsize
\caption{Performance comparisons of segmentation (above) and Detection (below) on the Kvasir-seg dataset with different networks. The notation $\uparrow$: higher is better.}
\label{tab2}
\centering
\resizebox{\columnwidth}{!}{%
\begin{tabular}{clccc}
\hline
\multicolumn{1}{l}{Model type}  & Model& \multicolumn{1}{l}{PA(\%)↑} &\multicolumn{1}{l}{meanIoU(\%)↑} & \multicolumn{1}{l}{Speed(fps)↑} \\ 
\hline
\multicolumn{1}{c|}{\multirow{3}{*}{\begin{tabular}[c]{@{}c@{}}
Single-task \end{tabular}}}  & U-net \cite{unet} & 83.37& 75.60 & 11 \\
\multicolumn{1}{c|}{} & Polyp-PVT \cite{polyppvt} & 91.49 & 86.40 & 17\\
\multicolumn{1}{c|}{} & Single-task baseline      & 90.95 & 86.24  & \textbf{41}  \\
\hline
\multicolumn{1}{c|}{\multirow{4}{*}{\begin{tabular}[c]{@{}c@{}}
Multi-task \end{tabular}}} & UOLO \cite{uolo}  & 83.41 & 75.48  & 9 \\
\multicolumn{1}{c|}{} & MULAN \cite{mulan} & 88.94 & 82.39 & 22\\
\multicolumn{1}{c|}{} & Multi-task Baseline & 90.88 & 85.73 & 36\\
\multicolumn{1}{c|}{} & \textbf{YOLO-Med(Ours)} & \textbf{97.32}& \textbf{88.64}& 31\\ \hline
\end{tabular}%
}

\resizebox{\columnwidth}{!}{%
\begin{tabular}{clccc}
\hline
\multicolumn{1}{l}{Model type}& Model  & \multicolumn{1}{l}{AP50(\%)↑} 
&\multicolumn{1}{l}{AP95(\%)↑} & \multicolumn{1}{l}{Speed(fps)↑} \\ 
\hline
\multicolumn{1}{c|}{\multirow{4}{*}{\begin{tabular}[c]{@{}c@{}}
Single-task\end{tabular}}}& Faster-RCNN \cite{faster}& 84.18 & 41.50 & 18 \\
\multicolumn{1}{c|}{} & RetinaNet \cite{retina}& 90.95 & 65.47 & 17 \\
\multicolumn{1}{c|}{} & YOLOv5s & 91.15 & 72.54 & \textbf{117}\\ 
\multicolumn{1}{c|}{} & Single-task baseline      & 91.31 & 72.66  & 47  \\
\hline
\multicolumn{1}{c|}{\multirow{4}{*}{\begin{tabular}[c]{@{}c@{}}
Multi-task\end{tabular}}} & UOLO \cite{uolo}  & 75.86 & 38.73 & 9\\
\multicolumn{1}{c|}{}& MULAN \cite{mulan}& 87.49& 53.40 & 22  \\
\multicolumn{1}{c|}{} & Multi-task Baseline & 89.73 & 67.11 &  36\\
\multicolumn{1}{c|}{}& \textbf{YOLO-Med(Ours)} & \textbf{94.72}& \textbf{73.02}& 31\\ \hline
\end{tabular}%
}

\vspace{-0.4cm}
\end{table}
\vspace{-5pt}
\subsection{Cross-scale task-interaction module}
\vspace{-3pt}
In this module, we combine features extracted by different decoders. As shown in Fig. \ref{MSTI}, we initially merge the outputs of the three different scales from the PAN network in the detection head to obtain a token sequence $\textbf{v}_{det}$,
\begin{equation}
\setlength{\abovedisplayskip}{3pt}
\setlength{\belowdisplayskip}{3pt}
  \textbf{v}_{det_{i}}' = \text{Reshape}(\text{Conv}(\textbf{x}_{det_i})),
\end{equation}
\begin{equation}
\setlength{\abovedisplayskip}{3pt}
\setlength{\belowdisplayskip}{3pt}
  \textbf{v}_{det} = \Vert\underset{i}{}(\textbf{v}_{det_{i}}'),
\end{equation}
where \emph{Conv} is used to standardize the channel number of features $\textbf{x}_{det_i}$ to 64. \emph{Reshape} is applied to flatten the feature $\textbf{v}_{det_i}' \in \mathbb{R}^{\frac{H}{s_i}{\times}\frac{W}{s_i}{\times}64}$ to a sequence $\textbf{v}_{det_i}' \in \mathbb{R}^{{n_i}{\times}64}$ with ${n_i} = \frac{H}{s_i}{\times}\frac{W}{s_i}$. Subsequently, we concatenate ($\Vert$) the three token sequences to obtain the final tocken sequence $\textbf{v}_{det} \in \mathbb{R}^{{(n_1 + n_2 + n_3)}{\times}64}$.

Similarly, we convert the feature map $\textbf{x}_{seg}$ from the segment head into a token sequence $\textbf{v}_{seg}$.
\begin{equation}
\setlength{\abovedisplayskip}{3pt}
\setlength{\belowdisplayskip}{3pt}
    \textbf{v}_{seg} = \text{Reshape}(\text{Conv}(\textbf{x}_{seg})),
\end{equation}
where \emph{Conv} is used to transform $\textbf{x}_{seg} \in \mathbb{R}^{\frac{H}{2}{\times}\frac{W}{2}{\times}32}$ to $\textbf{x}_{seg} \in \mathbb{R}^{\frac{H}{4}{\times}\frac{W}{4}{\times}64}$ . \emph{Reshape} is applied to flatten the feature $\textbf{x}_{seg} \in \mathbb{R}^{\frac{H}{4}{\times}\frac{W}{4}{\times}64}$ to a sequence $\textbf{v}_{seg} \in \mathbb{R}^{n_4 \times64}$ with ${n_4} = \frac{H}{4}{\times}\frac{W}{4}$.

We concatenate ($\Vert$) the two token sequences to obtain the final token sequence $\textbf{v}$.
\begin{equation}
\setlength{\abovedisplayskip}{3pt}
\setlength{\belowdisplayskip}{3pt}
    \textbf{v} = \Vert(\textbf{v}_{det}, \textbf{v}_{seg}),
\end{equation}
where $\textbf{v} \in \mathbb{R}^{n \times 64}$ with $n = \sum_{i=1}^{4} n_i$.

Next, we construct a Transformer Layer \cite{vit} with multi-head self attention (MHSA) as shown on the right side of Fig. \ref{MSTI}. 
\begin{equation}
\setlength{\abovedisplayskip}{3pt}
\setlength{\belowdisplayskip}{3pt}
    Q=\mathrm{MLP}(\textbf{v}), K=\mathrm{MLP}(\textbf{v}), V=\mathrm{MLP}(\textbf{v}),
\end{equation}
\begin{equation}
\setlength{\abovedisplayskip}{3pt}
\setlength{\belowdisplayskip}{3pt}
    \textbf{v}' = \mathrm{MHSA}(Q, K, V) + \textbf{v},
\end{equation}
\begin{equation}
\setlength{\abovedisplayskip}{3pt}
\setlength{\belowdisplayskip}{3pt}
    \hat{\textbf{v}} = \mathrm{MLP}(\mathrm{LN}(\textbf{v}')) + \textbf{v}',
\end{equation}
$\hat{\textbf{v}} \in \mathbb{R}^{n \times 64}$ with $n = \sum_{i=1}^{4} n_i $ is the cross-scale task-interaction feature. Here, \emph{LN} means LayerNorm and \emph{MLP} is the linear layer.

Conversely, we employ \emph{split} and \emph{reshape} operations to obtain feature maps with sizes consistent with the input features. We then restore the channel number using a convolutional layer.
\begin{equation}
\setlength{\abovedisplayskip}{3pt}
\setlength{\belowdisplayskip}{3pt}
    \widehat{\textbf{x}}_{det_i} = \text{Conv}(\text{Reshape}(\text{Split}(\hat{\textbf{v}}))_i) \in \mathbb{R}^{\frac{H}{s_i}{\times}\frac{W}{s_i}{\times}c_i},
\end{equation}
\begin{equation}
\setlength{\abovedisplayskip}{3pt}
\setlength{\belowdisplayskip}{1pt}
    \widehat{\textbf{x}}_{seg} = \text{Conv}(\text{Reshape}(\text{Split}(\hat{\textbf{v}}))_4) \in \mathbb{R}^{\frac{H}{2}{\times}\frac{W}{2}{\times}32},
\end{equation}

\subsection{Loss function}
The object detection loss ($\mathcal{L}_{\text{det}}$) comprises a weighted sum of the classification loss ($\mathcal{L}_{\text{class}}$), object loss ($\mathcal{L}_{\text{obj}}$), and bounding box loss ($\mathcal{L}_{\text{box}}$). As for segmentation loss ($\mathcal{L}_{\text{seg}}$), we utilize cross-entropy loss with logits ($\mathcal{L}_{\text{ce}}$). The global loss ($\mathcal{L}_{global}$) is as follows,
\begin{equation}
\setlength{\abovedisplayskip}{3pt}
\setlength{\belowdisplayskip}{3pt}
    \mathcal{L}_{\text{global}} = \beta_1 (\alpha_1 \mathcal{L}_{\text{class}} + \alpha_2 \mathcal{L}_{\text{obj}} + \alpha_3 \mathcal{L}_{\text{box}}) + \beta_2 \mathcal{L}_{\text{ce}},
\end{equation}
where $\alpha_1,\alpha_2,\alpha_3$ are uniformly set to $\frac{1}{3}$, and we set the weights of the detection loss and segmentation loss to be the same, with $\beta_1 = \beta_2 = \frac{1}{2}$. Both $\mathcal{L}_{\text{class}}$ and $\mathcal{L}_{\text{obj}}$ are implemented as focal loss \cite{focal}. Additionally, we employ the Localization Complete Intersection over Union $(\mathcal{L}_{CIoU})$ metric \cite{ciou} for $\mathcal{L}_{\text{box}}$.   

\begin{table}[t]
\scriptsize
\caption{Performance comparisons of segmentation (above) and detection (below) on our private dataset with different networks. The notation $\uparrow$: higher is better.}
\label{tab1}
\centering
\resizebox{\columnwidth}{!}{%
\begin{tabular}{clccc}
\hline
\multicolumn{1}{l}{Model type}& Model& \multicolumn{1}{l}{PA(\%)↑} 
&\multicolumn{1}{l}{meanIoU(\%)↑} & \multicolumn{1}{l}{Speed(fps)↑} \\ 
\hline
\multicolumn{1}{c|}{\multirow{3}{*}{\begin{tabular}[c]{@{}c@{}}
Single\\ model\end{tabular}}}& U-net \cite{unet}& 78.24& 57.94 & 10\\
\multicolumn{1}{c|}{} & Polyp-PVT \cite{polyppvt}& \textbf{90.60} & 71.12 & 16\\
\multicolumn{1}{c|}{} & Single-task Baseline & 86.39 & 70.64 & \textbf{37}\\
\hline
\multicolumn{1}{c|}{\multirow{3}{*}{\begin{tabular}[c]{@{}c@{}}
Multi-task\\ model\end{tabular}}} & UOLO \cite{uolo}& 80.41 & 63.28 & 8 \\
\multicolumn{1}{c|}{}& MULAN \cite{mulan} & 89.76 & 71.05& 19 \\
\multicolumn{1}{c|}{}& Multi-task Baseline& 83.24 & 68.97& 33 \\
\multicolumn{1}{c|}{}& \textbf{YOLO-Med(Ours)} & 89.96& \textbf{71.82}& 29\\ 
\hline
\end{tabular}%
}
\resizebox{\columnwidth}{!}{%
\begin{tabular}{clccc}
\hline
\multicolumn{1}{l}{Model type}& Model& \multicolumn{1}{l}{AP50(\%)↑} & \multicolumn{1}{l}{AP95(\%)↑} & \multicolumn{1}{l}{Speed(fps)↑} \\ 
\hline
\multicolumn{1}{c|}{\multirow{4}{*}{\begin{tabular}[c]{@{}c@{}}
Single\\ model\end{tabular}}}& Faster-RCNN \cite{faster} & 74.13& 20.61& 14\\
\multicolumn{1}{c|}{} & RetinaNet \cite{retina}& 76.80 & 29.63 & 14 \\
\multicolumn{1}{c|}{} & YOLOv5s & 76.66 & 33.85& \textbf{100}\\ 
\multicolumn{1}{c|}{} & Single-task Baseline& 77.21 & 33.44 & 42 \\
\hline
\multicolumn{1}{c|}{\multirow{4}{*}{\begin{tabular}[c]{@{}c@{}}
Multi-task\\ model\end{tabular}}} & UOLO \cite{uolo}& 60.85& 21.26 & 8\\
\multicolumn{1}{c|}{}& MULAN \cite{mulan} & 77.40 & 30.66 & 19\\
\multicolumn{1}{c|}{}& Multi-task Baseline& 76.30 & 33.21 & 33 \\
\multicolumn{1}{c|}{}& \textbf{YOLO-Med(Ours)} & \textbf{78.56}&\textbf{35.43}& 29\\ \hline
\end{tabular}%
}
\vspace{-0.4cm}
\end{table}
\vspace{-5pt}
\section{Experiments}
\label{sec:exp}
\vspace{-3pt}
\subsection{Experimental Setting}
\vspace{-3pt}
\textbf{Datasets.} We conduct training using two datasets: Kvasir-seg \cite{Kvasir}, a publicly available dataset, and a novel private biomedical dataset \cite{dataset}. Kvasir-seg comprises 1000 gastrointestinal disease images, each meticulously labeled for semantic segmentation and object detection. 
In addition, in collaboration with \textit{Shanghai General Hospital}, we create a novel private biomedical dataset \cite{dataset}. This dataset consists of images obtained through magnifying endoscopy with narrow-band imaging (ME-NBI) and includes annotations for both detection bounding boxes and polygon segmentation of gastric neoplastic lesions. It encompasses 3757 images collected from 392 patients, with annotations reviewed and verified by at least two experts.
For both datasets, we adopt a split of 70\% for training, 15\% for validation, and 15\% for testing.

\noindent \textbf{Implementation details.} Our network is trained using Stochastic Gradient Descent (SGD) optimization algorithm with a learning rate of $1\times10^{-2}$, weight decay of $5\times10^{-4}$ and momentum of 0.937. Additionally, we employ the Cosine Annealing with Warm Restarts learning rate scheduling strategy, in which the first three epochs server as warm-up epochs with a reduced learning rate. To initiate the training, we utilize a pre-trained model from the COCO dataset. All experiments are conducted on a single NVIDIA GeForce 2080Ti GPU.

\noindent \textbf{Baseline models and Metrics.} We comprehensively evaluate our network by comparing it with various biomedical multi-task networks, as well as networks specialized in either object detection or semantic segmentation tasks. For object detection, we consider high-performing models from recent years in the biomedical image domain, including RetinaNet \cite{retina}, as well as iconic models like Faster-RCNN \cite{faster} and YOLOv5s, representing two-stage and one-stage networks respectively. In addition, we include a comparison with our single-task baseline model, which consists of only the encoder and the detection decoder. We evaluate our model's detection accuracy using mean Average Precision at 50\% IoU (mAP50) and mean Average Precision at 95\% IoU (mAP95) as metrics. Regarding the semantic segmentation task, our comparisons encompass classic architecture U-net \cite{unet}, and Polyp-PVT \cite{polyppvt} which utilizes transformer modules to enhance accuracy. We also include a comparison with the single-task baseline which comprises only the encoder and the segmentation decoder. We evaluate our model's segmentation accuracy using Pixel Accuracy (PA) and mean Intersection over Union (meanIoU) as metrics. In the realm of multi-task networks, we compare our approach to the traditional UOLO \cite{uolo}, the latest MULAN \cite{mulan}, and the multi-task baseline of our model which includes only the encoder and two decoders. In addition to these horizontal comparisons with common networks, we conduct ablation experiments to investigate the impact of different modules within YOLO-Med, providing a detailed study of the network's components.

\begin{table}[t]
\caption{Ablation studies and analysis on Kvasir-seg (left) and our private dataset (right). Decoupled head (DH), Cross-Scale Task-Interaction (CSTI) module are the parts of our model. The notation $\uparrow$: higher is better. The w/ indicates ”with”.}
\label{tab3}
\resizebox{\columnwidth}{!}{%
\begin{tabular}{@{}lccccc@{}}
\toprule
\multicolumn{1}{c}{Models}      & \multicolumn{1}{l}{AP50(\%)↑} & \multicolumn{1}{l}{AP95(\%)↑} & \multicolumn{1}{l}{PA(\%)↑} & \multicolumn{1}{l}{meanIoU(\%)↑} & \multicolumn{1}{l}{Speed(fps)↑} \\ \midrule
\multicolumn{1}{l|}{Baseline}   & 89.73 & 67.11 & 90.88& 85.73& \textbf{36}\\
\multicolumn{1}{l|}{w/ DH}      & 92.01 & 72.98 & 91.59& 86.50& 34\\
\multicolumn{1}{l|}{w/ CSTI}    & 91.75 & 70.80 & 94.21& 88.43& 32\\
\multicolumn{1}{l|}{w/ DH+CSTI} & \textbf{93.78} & \textbf{73.02} & \textbf{94.32}& \textbf{88.56}& 31 \\\bottomrule
\end{tabular}%
}
\vspace{-0.5cm}
\end{table}

\vspace{-0.3cm}
\subsection{Experimental results}
\noindent \textbf{Object detection results}
In the evaluation on the public dataset Kvasir-seg \cite{Kvasir}, as presented in Table \ref{tab2}, our model outperforms single-task networks such as Faster-RCNN \cite{faster}, RetinaNet \cite{retina}, YOLOv5s and our single-task baseline, as well as all three multi-task networks in terms of detection accuracy. Notably, our model demonstrates impressive real-time performance compared to other works, with only YOLOv5s surpassing it. However, it's important to note that YOLOv5s lacks a segmentation decoder and a cross-scale task-interaction module. In the comparison between YOLO-Med and our single-task multi-task baseline models, all metrics indicate a higher level of object detection accuracy, with only a minimal 
decrease in inference speed. Similar results are observed on the private dataset as illustrated in Table \ref{tab1}.

\noindent \textbf{Semantic segmentation results}
In the evaluation on the public dataset Kvasir-seg, as shown in Table \ref{tab2}, our network outperforms all three multi-task networks and three single-task networks. Furthermore, our network exhibit significantly superior real-time performance compared to both single-task or multi-task networks from other works, such as U-net \cite{unet}, Polyp-PVT \cite{polyppvt}, UOLO \cite{uolo} and MULAN \cite{mulan}. When compared to our baseline models, all metrics surpass them, with an acceptable decrease in inference speed to enhance the segmentation accuracy. Similar results are observed on the large private dataset as shown in Table \ref{tab1}.

\begin{figure}[t]
\begin{center}
\setlength{\belowdisplayskip}{0pt}
\setlength{\abovecaptionskip}{1cm}
\includegraphics[width=0.95\linewidth]{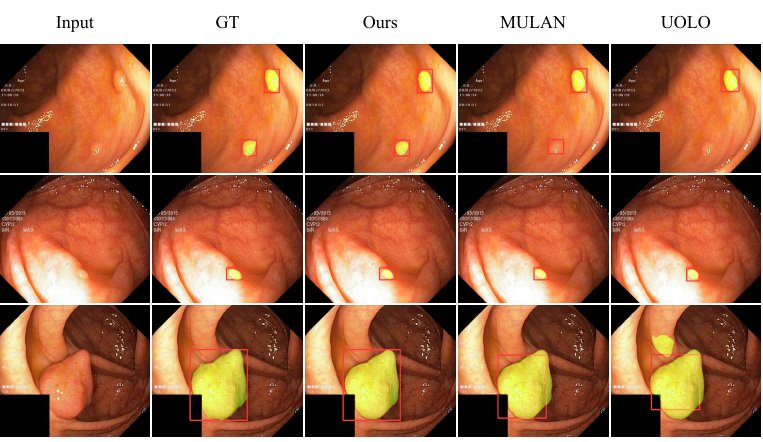}
\end{center}
\vspace{-0.5cm}
\caption{Qualitative comparison with two multi-task networks MULAN \cite{mulan} and UOLO \cite{uolo} on Kvasir-seg \cite{Kvasir}. The detection and segmentation results are shown in the same figure.}
\label{qualitative}
\vspace{-0.3cm}
\end{figure}

As depicted in Fig. \ref{qualitative}, we conduct a qualitative performance analysis by comparing our network with UOLO \cite{uolo} and MULAN \cite{mulan} on Kvasir-seg \cite{Kvasir}.  Our network produces more accurate predictions for detection and segmentation, whether it involves multiple small objects (top), single small object (middle) and single huge object (bottom).

\vspace{-0.3cm}
\subsection{Ablation studies}
In this section, we conduct four experiments, starting with a baseline and then introducing the Decoupled Head (DH) and the Cross-Scale Task-Interaction Module (CSTI) separately. We also evaluate a complete version that incorporates both modules. As presented in Table \ref{tab3}, from an accuracy perspective, the CSTI module has the most substantial positive impact on the segmentation task, with the network using only the CSTI module performing nearly as well as the complete version with both CSTI and DH modules. In contrast, the DH module is not able to bring huge improvements. The combined use of CSTI and DH yields the most substantial improvements. Regarding the detection task, the CSTI module alone brings noticeable improvements, while the DH module has a greater impact. Ultimately, the complete version with both CSTI and DH modules achieves the highest accuracy improvement. On the other hand, considering inference speed, utilizing the CSTI module alone leads to a 4 fps reduction compared to the baseline model, while employing only the DH module results in a 2 fps decrease due to its fewer parameters.

\begin{figure}[t]
\begin{minipage}[b]{0.5\linewidth}
  \centering
  \centerline{\includegraphics[width=3.5cm]{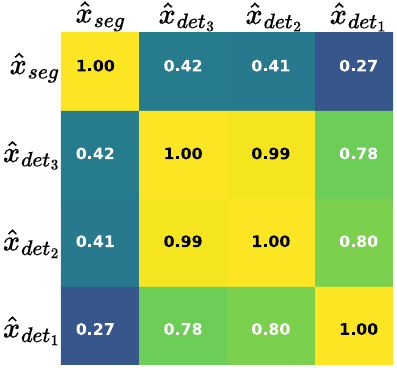}}
  \centerline{(a)}
\end{minipage}
\hfill
\begin{minipage}[b]{0.5\linewidth}
  \centering
  \centerline{\includegraphics[width=3.5cm]{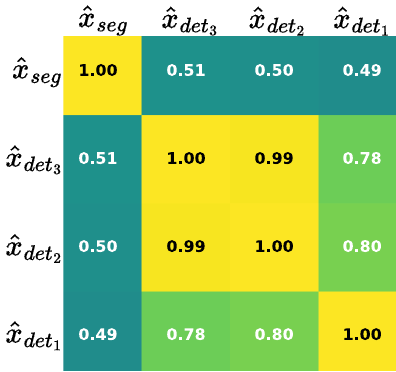}}
  \centerline{(b)}
\end{minipage}
\caption{Example correlation maps for the 4 outputs of the CSTI module. (a) depicts the correlation pattern for detecting and segmenting small objects, while (b) illustrates the scenario for large objects.}
\vspace{-0.5cm}
\label{abla}
\end{figure}

To enhance the CSTI module's effectiveness, as depicted in Fig. \ref{abla}, we conduct an analysis of correlations among its four outputs. Panel (a) presents results for detecting and segmenting small objects, while (b) for large objects. A comparison between them reveals correlations among the outputs of the detection and segmentation tasks. Notably, correlations within the detection task across different scales are consistently stronger than those between detection and segmentation tasks. However, the correlation between detection and segmentation tasks varies with object size. For small objects, the correlation between $\hat{x}_{seg}$ and $\hat{x}_{det_{1}}$ is only 0.27, whereas for large objects, this value increases to 0.49. These observations suggest that the CSTI module can dynamically adapt task relationships, effectively conveying information and ultimately enhancing overall performance.


\vspace{-9pt}


\section{Conclusion}
\vspace{-3pt}
In this paper, we present YOLO-Med, an efficient end-to-end multi-task network specifically designed to address both object detection and semantic segmentation tasks for biomedical image analysis. Our model excels in performance on two datasets: Kavarsir-seg and a private dataset. It not only achieves high accuracy in both tasks but also maintains real-time inference speed. Additionally, we validate the effectiveness of the proposed cross-scale task-interaction module, underscoring the value of cross-scale inter-task information fusion in the biomedical domain. This research carries significant implications for advancing future studies in the field of biomedical multi-task learning.


\vfill\pagebreak

\bibliographystyle{IEEEbib}
\bibliography{strings}

\end{document}